\documentclass[journal]{vgtc}   
\usepackage{booktabs}
\usepackage{tabularx}
\usepackage{array}
\usepackage{listings}
\usepackage{xcolor}
\usepackage[most]{tcolorbox}

\onlineid{1358}

\title{Food Critic: An Agentic Visualization System for Auditable Restaurant-Safety Reasoning}

\author{
\authororcid{Arleth Salinas}{0000-0003-1804-3511},
\authororcid{Ishrat Jahan Eliza}{0000-0003-3087-5951},
\authororcid{Valerio Pascucci}{0000-0002-8877-2042}
}

\authorfooter{
  \item
  	Arleth Salinas, Ishrat Jahan Eliza, and Valerio Pascucci are with the University of Utah. E-mail: [arleth.salinas, ishratjahan.eliza]@utah.edu.
}

\abstract{%
  Generative AI agents are increasingly being embedded into visualization systems that can search data, invoke tools, summarize evidence, and produce recommendations. However, when agents rely solely on their own knowledge, the visualization problem is not only whether the agent provides a useful answer, but also whether users can inspect how that answer is grounded. We present Food Critic, an agentic visualization system for diners' restaurant decision support that integrates public health inspection records with crowd review data. The system grounds agents under three conditions: a database-only ReAct agent, a retrieval-augmented agent with access to food-safety regulations, and an ontology-guided agent that produces structured hypotheses and typed evidence. Food Critic externalizes the agent’s reasoning within these groundings as inspectable visual evidence. We evaluate LLM models with several use cases under each grounding of our agentic system. Our work positions restaurant choice as a concrete case for studying human-agent collaboration, visual provenance of agent reasoning, and the design of visual interfaces for grounded GenAI decision support.
}


\teaser{
  \centering
  \includegraphics[width=\linewidth, alt={A view of a city with buildings peeking out of the clouds.}]{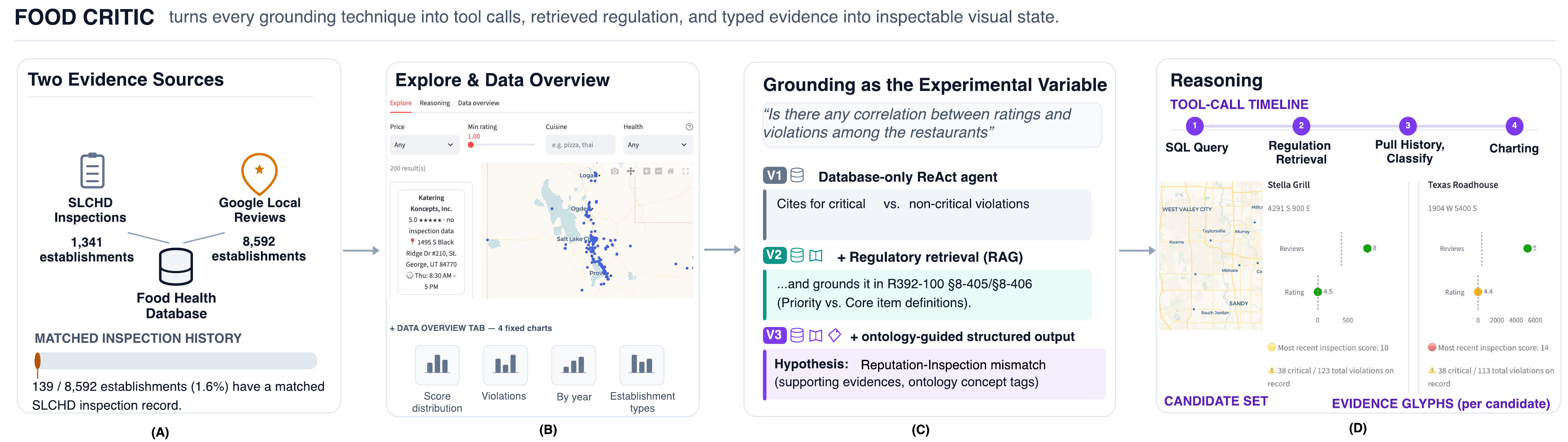}
  \caption{\textbf{Food Critic externalizes grounded LLM-agent reasoning as visual provenance.} Across three grounding conditions, the system combines restaurant inspection records, crowd reviews, regulatory retrieval, and ontology-guided evidence. The Reasoning view turns the agent's tool calls, candidate-set changes, retrieved regulations, and typed evidence into a linked visual state, allowing users to audit claims rather than accept a fluent narrative.}
  \label{fig:teaser}
}

\renewcommand{\manuscriptnotetxt}{}

\newcommand{\accslot}[1]{\makebox[0.62em]{#1}}
\newcommand{\accC}{\accslot{$\bullet$}}
\newcommand{\accW}{\accslot{$\circ$}}
\newcommand{\accF}{\accslot{\scriptsize$\times$}}
\newcommand{\accN}{\accslot{}}
\graphicspath{{figs/}{figures/}{pictures/}{images/}{./}} 

\usepackage{tabu}                      
\usepackage{booktabs}                  
\usepackage{lipsum}                    
\usepackage{mwe}                       

\usepackage{mathptmx}                  

\begin{document}


\firstsection{Introduction}

\maketitle

Generative AI agents are beginning to change the role of visualization systems. Instead of only helping people encode, filter, and inspect data, emerging systems allow agents to search databases, invoke analytic tools, generate charts, interpret results, and participate in sensemaking workflows. This shift creates a central challenge for visualization research: when an agent performs part of the analysis, how should a human user inspect, steer, and contest the agent’s reasoning? Recent work on agentic visualization has begun to identify design patterns for embedding agents into visual analytics systems while preserving human agency~\cite{dhanoa2025, liu2024ava, wang2024formulator}. However, many agentic systems still expose reasoning primarily as natural language, leaving users with limited support for tracing which data records, tool calls, regulatory facts, or assumptions shaped a response.

We study this problem in the context of restaurant decision support. Choosing a restaurant is an ill-defined decision: there is no single correct answer, no fixed objective function, and no stable weighting of criteria across users~\cite{dare2026, dimara2021}. A diner may care about distance, cuisine, rating, price, recent reviews, accessibility, or inspection history, and these criteria may conflict. Public health inspection records add another layer of complexity. They are authoritative, but often difficult to discover at the moment of choice; crowd platforms such as Google Maps and Yelp are visible and familiar, but their star ratings and review text do not reliably substitute for sanitation evidence. Prior work finds that review and inspection signals may align only in limited settings, such as chain establishments~\cite{yelpvsinspection}, and that tract-level review-derived illness signals may show minimal correlation with inspection outcomes~\cite{nyctract}. Thus, the decision problem is not simply to compute a better restaurant score, but to help users see what different evidence sources do and do not support.

An LLM agent can reconcile heterogeneous data sources on demand, retrieve regulatory context, and adapt its explanation to a user’s question. However, the same agent may also produce plausible yet unsupported judgments, especially when inspection coverage is sparse or when review sentiment is mistaken for evidence of safety. In public health contexts, retrieval grounding can reduce hallucination~\cite{megarag}, but grounding alone is not enough if users cannot inspect how it was used. The visualization question, therefore, is how to make an agent’s reasoning visible as structured, auditable evidence.

We present \textbf{Food Critic}, an agentic visualization system that pairs a database of government inspection records and crowd review data with an LLM agent. The system treats domain grounding as the main experimental variable. Across three conditions, the agent has access to the same core database tools but differs in whether it also receives regulatory retrieval and ontology-guided structured output. Our research questions are: 



\textbf{RQ1}: How does domain grounding change an LLM agent’s answers compared with a database-only reasoning and acting (ReAct) agent?

\textbf{RQ2}: How can an agent’s reasoning be externalized as an inspectable visual state, including tool calls, candidate sets, regulatory evidence, and typed support or contradiction?


Our contributions are: 1) A visual reasoning interface that externalizes an LLM agent's tool-call trace, candidate-set changes, retrieved regulations, and ontology-typed evidence as linked visual encodings. 2) Food Critic, an agentic visualization system for restaurant decision support that integrates public inspection records and crowd review data into a shared visual analytics workflow. 3) A grounding design for agentic VIS, combining regulatory retrieval with an ontology-guided hypothesis and evidence schema that makes evidence auditable claim-by-claim. 4) A controlled evaluation design that compares database-only, RAG-grounded, and ontology-guided agent conditions across prompt categories where grounding should or should not affect the answer.

\section{Related Work}
\textbf{Food inspection and review data as a visual analytics substrate.} A substantial body of work uses online restaurant reviews to infer or predict food inspection outcomes. Kang et al. first showed that Yelp review text contains latent sanitation signals~\cite{kang2013}, and subsequent work reported strong classification results on regional corpora using keyword, topic-model, and logistic-regression features~\cite{nguyen2016, schomberg2016}. The City of Boston operationalized this idea through an open competition that used review-derived signals to help prioritize inspections~\cite{drivendata2015}. A related public-health surveillance literature treats online reviews as a channel for detecting foodborne illness ~\cite{nsoesie2014}.


\textbf{Agentic visualization and reasoning provenance.} Food Critic is aligned with recent work on agentic visualization. Dhanoa et al.~\cite{dhanoa2025} reinterpret visualization systems through an agentic lens and identify design patterns for agent roles, communication, and coordination. Two patterns are particularly relevant: the Analyst role, in which an agent helps guide sensemaking over data, and the Provenance Log, which preserves an auditable record of sources, transformations, and actions. Food Critic uses these patterns in a domain decision-support setting: the agent analyzes restaurant evidence, while the Reasoning view turns its tool calls, retrieved evidence, and hypothesis classifications into a visual state. 


Current LLM-driven visualization systems can generate charts or visualization specifications from natural-language intent~\cite{wang2024formulator, wang2025formulator2,narechania2021, tian2024, vaithilingam2024, NL2ani} or provide conversational guidance for analysis and storytelling workflows~\cite{srinivasan2021, weng2025, wang2025jupybara}. Autonomous visualization agents have also been demonstrated in scientific and analytic settings~\cite{liu2024ava, liu2025paraview, yang2024matplotagent}, and multi-agent pipelines now support dashboard authoring, narrative generation, and data video production~\cite{dashchat2025, shen2024, fu2025, wolter2025, zhao2024}. Ye et al.~\cite{ye2024} survey this broader space of generative AI for visualization.

Food Critic differs from systems that primarily generate visual artifacts. Its central visual object is the agent’s reasoning process itself. A smaller body of work has begun to visualize agent behavior, including on-the-fly monitoring of conversational agent code~\cite{xie2024} and post hoc analysis of LLM-agent behavior~\cite{lu2025}. Food Critic extends this by making the agent’s intermediate actions evidence structures that end users can inspect. The Reasoning view treats tool calls and ontology-typed evidence as first-class visual marks.

\textbf{Grounding mechanisms for auditable agentic VIS.} Food Critic also relates to work on structured grounding for AI explanations. Ontologies have been used to constrain and explain recommendations in the food domain~\cite{foodontology}, and structured visual scaffolds can improve the interpretability of AI reasoning in high-stakes domains such as medical diagnosis~\cite{diagrammatization2025}. Retrieval-augmented generation is now a common method for reducing factual drift in domain applications~\cite{gao2024, ragreview}, with demonstrated benefits in public health question answering~\cite{megarag} and structured domain planning~\cite{epiplanagent}. At the same time, evidence from legal research tools indicates that retrieval reduces, but does not eliminate, unsupported assertions~\cite{magesh2024}.

In Food Critic, the regulatory retrieval provides domain-specific evidence about why a condition may constitute a health violation, while the ontology provides a vocabulary for how evidence should be used. This combination allows the system to externalize reasoning visually: users can see when the agent used the database, when it retrieved regulatory text, which ontology concept an evidence item supports, and whether a claim rests on sufficient inspection coverage. 

\textbf{Visualization for ill-defined decisions.} Dimara and Stasko~\cite{dimara2021} argue that decision-making tasks are often underspecified in visualization research and that few visualization tools directly support the preferential-choice step itself~\cite{oral2023}. Restaurant selection sits within this class of decision-making tasks. Such decisions are vulnerable to framing effects and cognitive bias~\cite{dimara2018, padilla2018}. These concerns are especially relevant for agentic systems, where a fluent recommendation may make a subjective judgment appear objective. Horvitz’s mixed-initiative interaction principles~\cite{horvitz1999}, Heer's account of agency plus automation~\cite{heer2019}, and Shneiderman’s human-centered AI framing~\cite{shneiderman2022} all argue that useful automation should preserve human control. 
Food Critic uses visualization to make the agent’s evidence, uncertainty, and assumptions inspectable so that users can decide which claims to trust.

\section{Food Critic System}
\label{sec:system}
Food Critic is designed for a diner in Salt Lake County who wants to weigh a restaurant's inspection history alongside its reviews, and who needs to judge whether an agent's recommendation is actually supported by evidence. Our visual state design goals are described in Table~\ref{tab:states}.
Using Food Critic, the user can browse restaurants directly, ask the agent a question, and inspect how the agent's answer was produced.
In subsequent sections, we describe the data source (\cref{sec:data}; \cref{fig:teaser}(A)), the agent and the grounding conditions we compare (\cref{sec:grounding}; \cref{fig:teaser}(C)), and the interface (\cref{sec:interface}; \cref{fig:teaser}(B) \& (D)).


\subsection{Data}
\label{sec:data}

Food Critic merges two public sources (\cref{fig:teaser}A). Salt Lake County Health Department (SLCHD) inspection records~\cite{slchd} provide inspection dates, scores, and cited critical and non-critical violations for 1{,}341 establishment listings. We constructed a Utah subset from the Google Local dataset~\cite{googlelocal}. It provides metadata, categories, hours, ratings, and review text for 8{,}592 food-related places. We match SLCHD listings to Google Local places by name and address similarity, confirmed by geographic distance after geocoding SLCHD addresses with the U.S. Census batch geocoder~\cite{us_census_geocoder} (OpenStreetMap as a fallback~\cite{openstreetmap}), and load both sources into a normalized DuckDB database. Only 139 of the 8{,}592 places (1.6\%) have a matched inspection history.

\subsection{Agent and Grounding Conditions}
\label{sec:grounding}

The agent is an LLM running a ReAct-style tool-calling loop~\cite{yao2023react} over the database (\cref{fig:teaser}B). It always has the same core tools: schema lookup, read-only SQL, a structured establishment search, per-restaurant inspection history, and chart rendering. The establishment search is the agent's main way to find restaurants, and its results are what the interface shows on the map. We compare three grounding conditions, each adding one layer to the previous:

\begin{itemize}
  \item \textbf{V1 (database only):} the core tools.
  \item \textbf{V2 (+ regulatory retrieval):} a retrieval tool over Utah Administrative Rule R392-100 (Food Service Sanitation)~\cite{r392100} and SLCHD's inspection-process documentation~\cite{slchd}, chunked and embedded in a vector index. Because retrieval is a tool call, every use of the regulation appears in the trace.
  \item \textbf{V3 (+ ontology-guided structured output):} a hand-authored ontology defines nine restaurant-safety hypotheses (e.g., \emph{health risk concern}, \emph{no inspection data available}, \emph{well reviewed}), the evidence each requires, a mapping from raw violation text to violation types, and data-quality flags. V3 must return a structured response: a hypothesis label, a confidence, and supporting and undermining evidence items, each tagged with an ontology concept and tied to the tool call it came from.
\end{itemize}

V1--V3 are configurations of the same agent that we compare in \cref{sec:results}; they are not modes a user selects. The interface is the same for all three, except for the Reasoning view, which requires V3's structured output. Schema details, retrieval construction, ontology contents, and prompts are in the supplemental materials.

\subsection{Interface}
\label{sec:interface}

The interface separates three kinds of visual state. Each is shown in a specific place (\cref{tab:states}).

\begin{table}[t]
  \centering
  \caption{The three kinds of visual state in Food Critic and where each appears.}
  \label{tab:states}
  \small
  \begin{tabular}{@{}p{0.22\linewidth}p{0.36\linewidth}p{0.34\linewidth}@{}}
    \toprule
    State & Contents & Where it is shown \\
    \midrule
    Database state (G1) & Restaurants, reviews, and inspection records, computed directly from the database & Explore list and map while browsing; Data Overview tab \\
    \addlinespace
    Agent-generated analytic state & The agent's answer, its charts, and the restaurants its searches returned & Chat panel; the returned restaurants also replace the Explore list and map \\
    \addlinespace
    Agent-reasoning state (G2) & How the answer was produced: tool calls, candidate sets, and typed evidence & Reasoning view (V3 only) \\
    \bottomrule
  \end{tabular}
\end{table}

\textbf{Explore view and Data Overview (\cref{fig:teaser}(B)).} The Explore view pairs a results list, filterable by price, minimum rating, cuisine, and inspection-data availability, with a map. Each result card shows the rating, address, and hours, and states ``no inspection data'' when the restaurant has no matched record (G1). The view works without the agent (G1). When the user asks a question, the restaurants returned by the agent’s searches replace the list and map and are labeled with the question they answer. Users can clear the results to return to browsing. The answer itself, with any charts the agent generated, appears in the chat panel. The Data Overview tab contains four fixed charts computed directly from the database, independent of any agent call: inspection-score distribution, top violation categories, violations per year by severity, and establishment types.

\textbf{Reasoning view (\cref{fig:teaser}(D)).} The Reasoning view, available in V3, shows how the latest answer was produced (G2). It states the answer's hypothesis and confidence. A timeline scrubber steps through the agent's database searches; at each step, the map shows that search's results, so the user can see how the candidate set changed. Other steps, such as regulation retrievals, SQL queries, and violation classifications, are listed below the timeline. Evidence glyphs show one card per candidate restaurant: each criterion (rating, number of reviews, and distance when relevant) is plotted against a threshold taken from corpus percentiles and colored as good, warning, or critical, and the card states whether the restaurant has a matched inspection history. The candidates come from the agent's actual search results, not from the model's text, so a glyph never shows a restaurant the agent did not retrieve.

\section{Results and Discussion}
\label{sec:results}

In this section, we present our findings through exploratory usage and analysis of the Food Critic system.
\subsection{Evaluation Design}
\label{sec:eval-design}

We compare the three grounding conditions (\cref{sec:grounding}) using a fixed script of 18 tasks totaling 21 turns: 15 single-turn questions (Q1--Q15) and three two-turn follow-up threads (A1/A2, B1/B2, C1/C2), in which the second turn is asked in the same conversation as the first. We ran the script once per condition with five LLMs: Anthropic's Claude Sonnet 5, Google's Gemini 3.6 Flash, Groq-hosted Llama 3.3 70B, OpenAI's GPT-5.6 Terra, and Google's open-weight Gemma 4 26B run locally through Ollama.

\textbf{Task definition.} We wrote the questions to cover five kinds of tasks a diner might bring to the system. The set deliberately includes tasks where grounding should \emph{not} matter, so that the comparison does not favor added grounding by design:
\begin{itemize}
  \item \emph{Database lookup and ranking} (Q1, Q3, Q4, Q13, Q14, A1, B1): the core database tools should be sufficient.
  \item \emph{Visual analysis} (Q2, Q10--Q12, A2): tests the shared charting tools.
  \item \emph{Regulatory interpretation} (Q5, Q6, Q9): retrieval (V2, V3) should help, because scoring rules and violation definitions are not in the database.
  \item \emph{Restaurant-specific safety judgment} (Q7, Q8): the ontology's hypotheses (V3) target these.
  \item \emph{Preference-based recommendation} (Q15, B2, C1, C2): open-ended, with no single correct answer.
\end{itemize}
The 17 turns in the first four types are graded against ground-truth answers computed directly from the database; the four recommendation turns are assessed qualitatively. The full script and scoring rubric are in the supplemental materials.

Each run logs the full message trace, every tool call and its raw result, chart specifications, the final answer or structured response, and token usage. Because the conditions share the same database and tools, differences in output within one LLM can be attributed to the added grounding layers.

\subsection{Factual Question Accuracy}
\label{sec:results-accuracy}
Across 303 total runs, 275 produced a response that did not result in an LLM API error (90.8\%); see Section~\ref{sec:results-failures} for the breakdown of the 28 failures. 17 of the 21 questions/thread-turns have a single, ground-truth-verifiable answer; the remaining 4 are open-ended recommendations with no single correct answer. Each of the 17 was graded against a set of ground truth answers. Each LLM is run once per grounding condition, and each run is graded individually. Table~\ref{tab:accuracy-grid} shows the outcome of every run; accuracy percentages count only runs that returned an answer (that is, no API error).

\begin{table*}[t]
  \centering
  \caption{Factual-answer accuracy by question and model provider. Each cell shows one mark per grounding condition, in the order V1 (database only), V2 (+RAG), V3 (+ontology-guided reasoning): \accC{}~correct, \accW{}~incorrect, \accF{}~LLM API error (no answer to grade); an empty slot means the run was not performed. Accuracy columns report the percentage of correct answers among runs that returned an answer. Q15, B2, C1, and C2 are open-ended recommendations with no single correct answer and are not graded. Q1: most-violating restaurant; Q2: rating/violation correlation; Q3: most-reviewed + history; Q4: most-reviewed with a violation; Q5: what is an inspection score; Q6: critical vs.\ non-critical; Q7: In-N-Out safety; Q8: Stella Grill trend; Q9: cold holding definition; Q10: critical-violation map; Q11: score distribution/stats; Q12: 5-year critical trend; Q13: highest 3-star street; Q14: Thai vs.\ Korean; A1: matched-inspection count; A2: distribution/timeline; B1: highest-rated vs.\ most-reviewed.}
  \label{tab:accuracy-grid}
  \footnotesize
  \setlength{\tabcolsep}{2.2pt}
  \begin{tabular}{@{}l*{14}{c}@{\quad}*{3}{c}@{\quad}c@{\quad}*{3}{c}@{}}
    \toprule
    & \multicolumn{14}{c}{Single-turn questions}
    & \multicolumn{3}{c}{Threaded}
    & & \multicolumn{3}{c}{Accuracy (\%) by condition} \\
    \cmidrule(lr){2-15} \cmidrule(lr){16-18} \cmidrule(lr){20-22}
    Provider & Q1 & Q2 & Q3 & Q4 & Q5 & Q6 & Q7 & Q8 & Q9 & Q10 & Q11 & Q12 & Q13 & Q14 & A1 & A2 & B1 & Overall (\%) & V1 & V2 & V3 \\
    \midrule
    Claude & \accC\accC\accF & \accC\accC\accF & \accC\accC\accF & \accC\accC\accF & \accC\accC\accF & \accC\accC\accF & \accC\accC\accF & \accC\accC\accF & \accC\accC\accN & \accC\accC\accN & \accC\accC\accN & \accC\accC\accN & \accC\accC\accN & \accC\accC\accN & \accC\accC\accF & \accC\accC\accF & \accC\accW\accF & 97 & 100 & 94 & -- \\
    Gemini & \accC\accC\accC & \accC\accC\accC & \accC\accC\accC & \accC\accC\accC & \accC\accC\accC & \accC\accC\accC & \accC\accC\accC & \accC\accC\accC & \accC\accC\accC & \accC\accC\accC & \accC\accC\accC & \accC\accC\accC & \accC\accC\accC & \accC\accC\accC & \accC\accC\accC & \accC\accC\accC & \accC\accC\accC & 100 & 100 & 100 & 100 \\
    Llama 3.3 (Groq) & \accC\accC\accW & \accW\accF\accC & \accF\accC\accC & \accC\accC\accW & \accC\accF\accC & \accC\accC\accC & \accF\accC\accW & \accW\accW\accW & \accC\accC\accC & \accF\accW\accC & \accW\accC\accW & \accW\accC\accW & \accC\accW\accW & \accF\accF\accC & \accW\accC\accC & \accC\accC\accC & \accF\accF\accW & 62 & 58 & 77 & 53 \\
    GPT & \accC\accC\accC & \accC\accC\accC & \accC\accC\accC & \accC\accC\accC & \accC\accC\accC & \accW\accC\accC & \accC\accC\accC & \accC\accC\accC & \accC\accC\accC & \accC\accC\accC & \accC\accC\accC & \accC\accC\accC & \accC\accC\accC & \accN\accN\accN & \accW\accW\accC & \accC\accC\accC & \accC\accC\accC & 94 & 88 & 94 & 100 \\
    Gemma 4 (Ollama) & \accC\accC\accC & \accC\accW\accC & \accC\accC\accC & \accW\accC\accC & \accC\accC\accC & \accC\accC\accC & \accC\accW\accC & \accC\accC\accC & \accC\accC\accC & \accC\accC\accC & \accN\accN\accN & \accW\accW\accW & \accC\accW\accC & \accC\accC\accC & \accW\accW\accW & \accC\accC\accC & \accC\accC\accC & 79 & 81 & 69 & 88 \\
    \bottomrule
  \end{tabular}
\end{table*}

Gemini answered every attempted factual question correctly (51/51); Claude was correct on 97.1\% on the 2 conditions it could complete (every attempted V3 run failed, and the remaining V3 questions were not run; Section~\ref{sec:results-failures}). Groq was the weakest LLM at 61.9\%, failing entirely on Q8 (0/3: every Groq run either omitted the trend analysis, the question asked for, or asserted a recent worsening trend that contradicts the data). The locally run Gemma 4 reached 79.2\% (38/48). It missed Q12 in every condition (no chart for an explicit visual-analysis request, or an empty answer) and A1 in every condition, where it reported the 1{,}315 inspected SLCHD listings instead of the 139 establishments with a matched inspection history.



\subsubsection{A data-coverage limit, not a reasoning limit}
\label{sec:results-coverage}

Q1 (``which restaurant violated health codes most'') surfaced a data cleaning issue. The true top violator by inspection data has zero rows in the Google Local data; only 139 of 1{,}341 SLCHD-tracked establishments have a matching row there. That restaurant is therefore unreachable through the system's own tools regardless of model quality, and Table~\ref{tab:accuracy-grid}'s Q1 column is graded against the tool-reachable ground truth, not the countywide one. Two runs (Gemini V1 and Gemma V2) named the countywide top violator anyway: after the establishment search returned the tool-reachable answer, each wrote its own SQL query over the full inspection tables and reported that result. We grade both as correct, since each claim is supported by the run's own trace.

\subsection{Grounding and Reasoning Traceability}
\label{sec:results-grounding}
\subsubsection{Per-condition summary}
\label{sec:results-per-condition}

The right-hand columns of Table~\ref{tab:accuracy-grid} break the per-provider
totals out by grounding condition. Gemini was perfect on all three (17/17
each); grounding condition made no difference to its final accuracy. Claude
was near-perfect on V1/V2 (17/17, 16/17) but produced zero successful V3
runs to grade at all, a hard schema-complexity failure, not a reasoning gap
(Section~\ref{sec:results-failures}). GPT improved slightly across
conditions (14/16, 15/16, 16/16). Groq's accuracy was weak on every
condition that required producing a visualization alongside a numeric claim,
regardless of grounding. Its tool-call argument-formatting failures happen before the RAG or ontology context is
even used, so grounding cannot fix them (Section~\ref{sec:results-failures}). Gemma dipped on V2 (13/16, 11/16, 14/16); three of its five V2 misses were
runs that ended without a final answer rather than wrong answers.

\subsubsection{How domain grounding changes the answer}
\label{sec:results-rq1}

\begin{figure}[t]
\centering
\includegraphics[width=\linewidth]{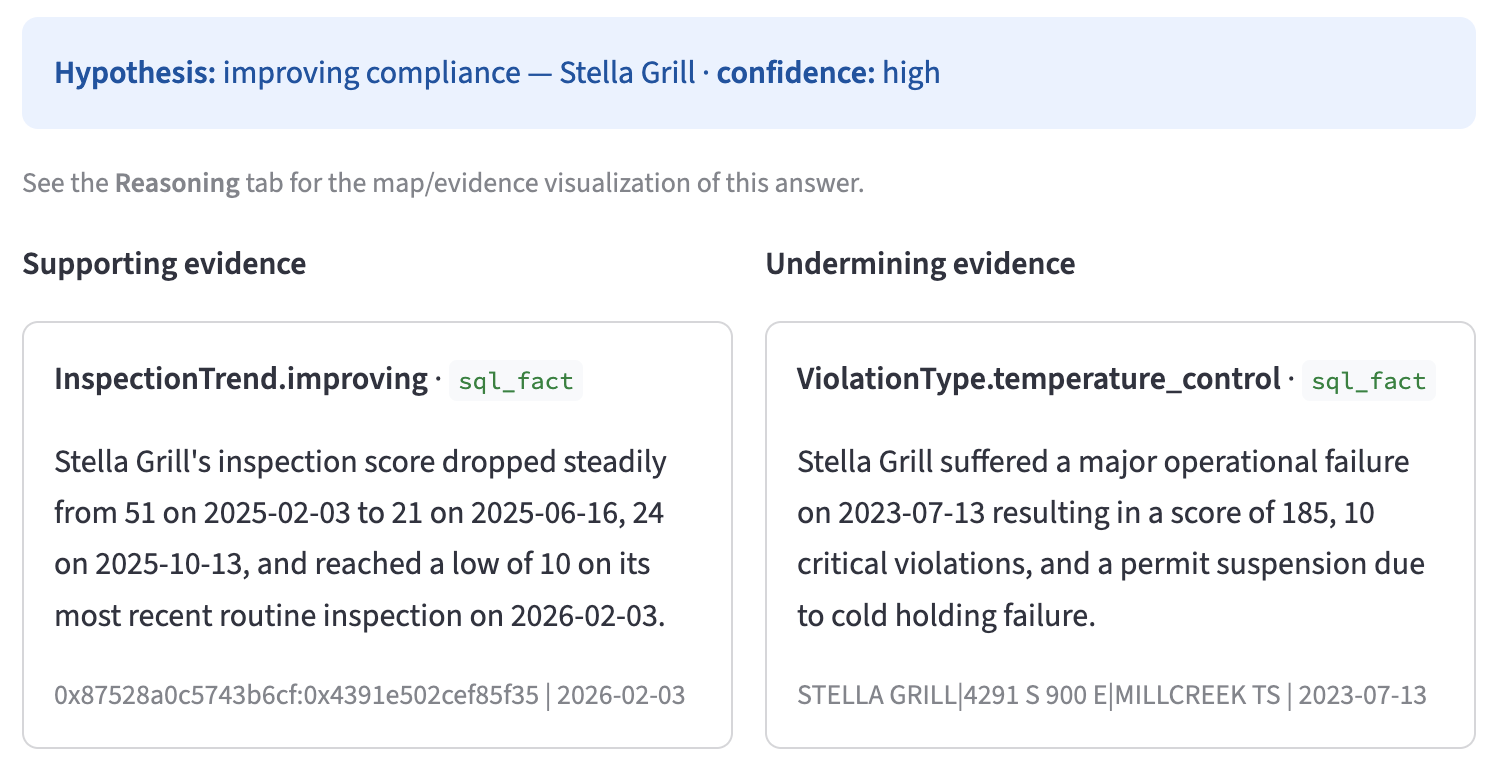}
\includegraphics[width=\linewidth]{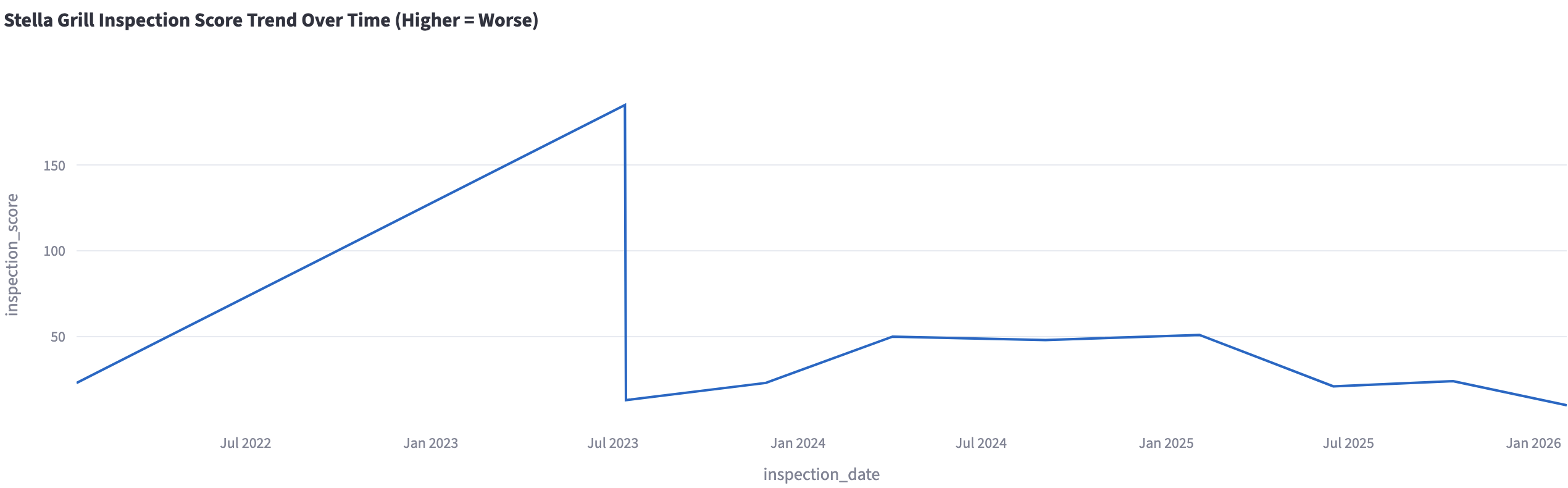}
\caption{Hypothesis and evidence provided by V3 using Gemini 3.6 Flash to the question "STELLA GRILL's inspection scores have fluctuated significantly over time. Should I be concerned? Answer with very detailed visual analytics."}
\label{fig:rq1-v1-v3}
\end{figure}

Figure~\ref{fig:rq1-v1-v3} shows the hypothesis and part of the evidence provided by V3 to support its answer to the question "STELLA
GRILL’s inspection scores have fluctuated significantly over time.
Should I be concerned? Answer with very detailed visual analytics." Gemini's aggregate accuracy
is identical across all three grounding conditions on this question
(Table~\ref{tab:accuracy-grid}). Grounding is not changing what the agent concludes here; what changes is how the conclusion is delivered: V1's answer is free-form prose with a chart and a markdown comparison table, with a claim "the establishment was issued a summary permit suspension" that has no pointer to which regulation authorizes that action or which data row supports it. V3's answer states identical facts but as a named hypothesis, a confidence score, and an evidence list, where every claim includes a source that one can trace back to a specific SQL row or regulatory section. In this case, domain grounding made the agent's reasoning inspectable.

\subsection{Failed Runs}
\label{sec:results-failures}

28 of 303 runs failed with an error at the LLM API call. The two failure modes are: Claude's V3 condition hitting
Anthropic's structured-output complexity limit when the ontology-guided
schema is combined with the full tool set (14 failures, all V3), and Groq
emitting malformed tool-call arguments (stringified booleans/integers where
the schema requires native JSON types) that fail server-side validation
before reaching our tool code (12 failures, spread across variants). Gemma had no API-level failures, but four of its runs completed without error and returned an empty final answer; we grade these as incorrect rather than count them as failures.

\section{Limitations, Future Work, and Lessons Learned}
Food Critic has several limitations. First, only a small fraction of Google Local establishments have matched SLCHD inspection histories. This limits what the agent can safely claim about food safety. Second, the evaluation is based on a scripted set of questions rather than a user study. Third, LLM behavior itself became a system-level constraint: structured-output complexity limits or malformed tool-call arguments.

For future work we plan to improve entity resolution and data coverage so that inspection evidence can be surfaced for more establishments and so that unmatched restaurants are represented with clearer provenance. We also plan to evaluate the Reasoning view with users, focusing on whether visual traces help people notice missing evidence, challenge unsupported recommendations, and compare conflicting review and inspection signals.

Food Critic shows that grounded agents need visual audit structures. Retrieval and ontology guidance can make an answer more domain-aware, but the user-facing contribution comes from externalizing how the answer was produced: which tools were called, which restaurants entered or left the candidate set, which regulatory passages were retrieved, and which evidence items supported or undermined the final claim. The Reasoning view shifts the visualization target from the restaurant data alone to the agent's reasoning over it. This distinction is important for agentic visualization systems because a fluent explanation can still hide missing evidence. In contrast, a visual trace can reveal where the agent's answer rests on database records, retrieved regulations, typed evidence, or abstention.
\section{Conclusion}
Food Critic demonstrates how an agentic visualization system can make grounded LLM reasoning visually inspectable. By comparing database-only, RAG-grounded, and ontology-guided agents over the same restaurant corpus, the system shows that grounding is not only a backend accuracy technique but also a visualization design problem. The key contribution is the transformation of agent reasoning into visual provenance: tool calls, candidate-set changes, retrieved regulation, and typed evidence become inspectable interface state. This case suggests that future agentic visualization systems should not only generate answers, but also expose the evidentiary structure that makes those outputs contestable.

\section*{Supplemental Materials}
The supplemental materials include a system walkthrough, annotated interface screenshots, the complete 21-question evaluation script, the three system prompts, charting policy text, tool-interface descriptions, a database schema summary, RAG construction details, ontology and hypothesis vocabulary, a run-log schema, and a scoring rubric. For the source data, we provide schema documentation, processing notes, descriptions of derived aggregates, and representative de-identified examples rather than raw input files. A link to our GitHub repository is available at \url{https://github.com/arlethzuri/Food-Critic}.






\bibliographystyle{abbrv-doi-hyperref}

\bibliography{template}


\lstdefinestyle{promptblock}{
  basicstyle=\ttfamily\scriptsize,
  breaklines=true,
  breakatwhitespace=false,
  columns=fullflexible,
  keepspaces=true,
  frame=single,
  framerule=0.3pt,
  xleftmargin=0.5em,
  xrightmargin=0.5em,
  aboveskip=0.75em,
  belowskip=0.75em
}

\newtcolorbox{promptbox}[1][]{
  enhanced,
  sharp corners,
  colback=black!5,      
  colframe=black,       
  coltitle=white,       
  fonttitle=\normalsize, 
  boxrule=0.5pt,        
  left=1em,             
  right=1em,
  top=0.75em,
  bottom=0.75em,
  title=#1              
}

\newcommand{\codepath}[1]{\texttt{\detokenize{#1}}}


\appendix

\section{Supplemental Materials Overview}
\label{app:overview}

This supplemental document provides additional details for Food Critic, an agentic visualization system for auditable restaurant-safety reasoning. The main paper introduces the system design and research questions; this supplement documents the artifacts needed to inspect the implementation and evaluation protocol in greater detail. A link to the GitHub repository of Food Critic is available here \url{https://github.com/arlethzuri/Food-Critic}. A link to a video demo of the system is available here \url{https://youtu.be/FNT6OvDdu5s}.

The supplement is organized as follows. Section~\ref{app:data} describes the review and inspection data sources, database construction, and schema. Section~\ref{app:rag} documents the regulatory retrieval corpus used by the RAG condition. Section~\ref{app:ontology} describes the ontology, hypothesis labels, violation-type mapping, and data-quality flags. Section~\ref{app:tools} lists the agent tools and their role in the visual interface. Section~\ref{app:prompts} includes the prompt text used to define the three grounding conditions. Section~\ref{app:questions} gives the complete 21-question evaluation script. Section~\ref{app:rubric} defines the scoring rubric and failure categories.

\section{Data Corpus and Integration}
\label{app:data}

\subsection{Source Data}

Food Critic uses the same review and inspection database for all three agent variants. All variants query the same DuckDB database, so the experimental comparison changes only the grounding available to the agent, not the underlying data.

The database is built from public inspection records and crowd-review data. The inspection records come from the Salt Lake County Health Department (SLCHD). The review records come from the Utah subset of the Google Local dataset. The source files used for the review and inspection portions of the database are summarized in Table~\ref{tab:data-sources}.

\begin{table*}[t]
\centering
\caption{
Source files used to construct the shared Food Critic review--inspection database.
All three agent variants query the same DuckDB database built from these files; the experimental conditions differ only in the grounding layers available to the agent.
}
\label{tab:data-sources}
\scriptsize
\begin{tabularx}{\textwidth}{p{0.18\textwidth} p{0.32\textwidth} X}
\toprule
\textbf{Table} & \textbf{Source CSV} & \textbf{Role} \\
\midrule
\texttt{Review} & \codepath{data/processed/review-Utah_food.csv} & Google Local review records. This file is loaded directly from the processed Google Local subset and is not produced by the inspection-cleaning notebook. \\
\texttt{Inspections} & \codepath{data/final/inspections.csv} & SLCHD inspection events, including inspection date, inspection type, establishment name, and inspection score. \\
\texttt{Violations} & \codepath{data/final/violations.csv} & Violation-grain inspection records. Each row represents a cited violation line associated with an inspection. \\
\texttt{Violation Codes} & \codepath{data/final/violation_codes.csv} & Violation-code lookup table, including code text, asterisk count, and critical-violation flag. \\
\texttt{Inspection Types} & \codepath{data/final/inspection_types.csv} & Inspection-type lookup table, including routine, follow-up, and other inspection categories. \\
\bottomrule
\end{tabularx}
\end{table*}


\subsection{Entity Resolution}

The database links SLCHD listings to Google Local establishments through an entity-resolution process based on establishment name, address, city, and geocoded location. A small fraction of Google Local food-tagged establishments have a matched SLCHD inspection history. Thus,  a restaurant may have rich review data but no matched inspection record. Food Critic therefore preserves unmatched establishments instead of filtering them out, and both the agent prompts and the visual interface are designed to represent missing inspection evidence explicitly.

\subsection{Geocoding}

Establishment addresses are geocoded using the U.S. Census batch geocoder, with OpenStreetMap Nominatim as a fallback. The resulting coordinates support spatial browsing, restaurant search, and candidate-set replay in the Reasoning view. Geocoding is used for visualization and proximity-oriented queries, not as a substitute for a verified public-health identifier.

\subsection{Database Schema}

The core database schema is summarized below. Arrows indicate foreign-key relationships used by the agent tools and interface.

\begin{lstlisting}[style=promptblock,caption={Core review and inspection schema.}]
review(
  user_id,
  name,
  "time",
  rating,
  text,
  resp,
  gmap_id -> establishments.gmap_id
)

inspection_types(
  inspection_type_id PRIMARY KEY,
  description
)

inspections(
  inspection_id PRIMARY KEY,
  slchd_establishment_key -> establishment_keys,
  inspection_date,
  inspection_type_id -> inspection_types.inspection_type_id,
  inspection_score
)

violation_codes(
  violation_code_id PRIMARY KEY,
  code_text,
  asterisk_count,
  critical
)

violations(
  violation_id PRIMARY KEY,
  inspection_id -> inspections.inspection_id,
  violation_code_id -> violation_codes.violation_code_id,
  violation_description,
  violation_occurrences,
  violation_cos,
  violation_phr
)
\end{lstlisting}

The \texttt{inspections} and \texttt{violations} tables jointly create the list of \texttt{establishments}, which represents the SLCHD listing.


Food Critic does not treat the integrated corpus as a complete ground-truth database. Instead, the corpus reflects a public-facing decision environment in which review data are often visible while inspection evidence is incomplete, difficult to locate, or unavailable for a matched establishment. Only 139 of 8,592 Google Local food-tagged establishments have a matched SLCHD inspection history. The remaining establishments are not assumed to be safe; they are represented as having no matched inspection evidence in this corpus.

This coverage asymmetry affects all three agent variants because all variants query the same database. It is also explicitly encoded in the V3 ontology through the hypothesis label. This label provides the agent with a structured way to refrain from making a safety judgment when inspection evidence is absent. It also gives the visualization a concrete state to render: lack of inspection evidence is not visually encoded as favorable evidence.

The system therefore distinguishes three cases that are easy to conflate in ordinary restaurant interfaces: an establishment with no matched inspection history, an establishment with inspection history and no cited violations for a given inspection, and an establishment with cited critical or non-critical violations. This distinction is necessary for visual reasoning over public-health evidence because a missing record and a clean record have different meanings.

\section{Regulatory Retrieval Corpus}
\label{app:rag}

The V2 and V3 agents use a regulatory retrieval corpus to ground explanations of food-safety rules, inspection scoring, and the meaning of violations. The raw regulatory source is Utah Administrative Rule R392-100, Food Service Sanitation Rule. The corpus also includes Salt Lake County's food-protection inspection-process page, which is fetched during ingestion.

The vector store is rebuilt by the RAG ingestion script and persisted to ChromaDB. The ingestion process converts PDF pages and the county inspection process page into text chunks using a recursive character splitter with a chunk size of 1000 and an overlap of 150. Chunks are embedded with the local transformers \texttt{all-MiniLM-L6-v2} model and stored in Chroma. This local embedding step avoids external API cost during ingestion.

Each retrieved chunk includes metadata fields such as source, source URL, page, and the best-effort nearest heading. The nearest heading is extracted by regular-expression matching of section-like identifiers, such as \texttt{1-101.10}, when available. At runtime, the retrieval corpus is made available to the agent via the \texttt{retrieve regulation} tool. This design makes regulatory grounding an observable tool action rather than a hidden prompt context.

The distinction matters for visualization. When the agent retrieves regulation text, the retrieval call appears in the tool trace and can be linked to the final claim. Thus, the Reasoning view can distinguish claims grounded in database records from claims grounded in regulatory text.

\section{Reasoning Ontology}
\label{app:ontology}

\subsection{Purpose}

The V3 condition adds a reasoning ontology that constrains the agent's answer into typed hypotheses and evidence items. The ontology serves two roles. First, it gives the agent domain vocabulary for cases where ordinary review-based recommendation language would be insufficient or misleading. Second, it gives the visualization structured objects to render. A free-text explanation can only be read as a narrative, but an ontology-guided answer can be decomposed into candidate restaurants, hypothesis labels, supporting evidence, undermining evidence, source references, and ontology concepts.

\subsection{Ontology Contents}

The ontology contains database-backed classes, derived concepts, hypothesis labels, and an evidence checklist used to guide final responses. The class list includes operational entities such as \texttt{Establishment}, \texttt{SLCHD Listing}, \texttt{Category}, \texttt{Establishment Category}, \texttt{Establishment Attribute}, \texttt{Operating Hours}, \texttt{Review}, \texttt{Inspection Type}, \texttt{Inspection}, \texttt{ViolationCode}, and \texttt{Violation}. It also includes derived or reasoning-oriented concepts such as \texttt{Violation Public Health Rationale}, \texttt{Violation Type}, \texttt{Inspection Trend}, \texttt{Hypothesis}, and \texttt{Evidence}.

The ontology also declares named relationships between classes, such as matching a Google Local establishment to an SLCHD listing, connecting inspections to violations, and mapping raw violation descriptions to normalized violation types. These relationships make the database joins explicit in the reasoning layer rather than leaving them implicit in SQL.

\subsection{Hypothesis Vocabulary}

The V3 agent must choose one hypothesis label from a fixed vocabulary, or choose \texttt{none} when the question is a plain lookup. The current vocabulary is: \texttt{safe bet}, \texttt{health risk concern}, \texttt{improving compliance}, \texttt{declining compliance}, \texttt{repeat critical violator}, \texttt{no inspection data available}, \texttt{well reviewed}, \texttt{reputation inspection mismatch}, \texttt{limited track record}, \texttt{none}.

Several labels are designed specifically for ambiguous or failure-prone evidence conditions. For example, \texttt{reputation inspection mismatch} field captures cases where review reputation and inspection history disagree. \texttt{limited track record} field captures cases where review volume is too small to support a strong reputation claim. \texttt{no inspection data available} captures the common case in which a restaurant has review data but no matched inspection record.

\subsection{Violation-Type Mapping}

The file \codepath{semantic/violation_map.csv} maps raw public-health-rationale strings from \texttt{violations.violation\_phr} to a smaller \texttt{ViolationType} vocabulary. The mapping is loaded by the ontology layer and exposed through the \texttt{classify\_violation} tool. The normalized violation types are:

\begin{itemize}
    \item \texttt{temperature control}
    \item \texttt{cross contamination}
    \item \texttt{hygiene practice}
    \item \texttt{pest presence}
    \item \texttt{facility maintenance}
    \item \texttt{documentation}
    \item \texttt{other}
\end{itemize}

This mapping implements the ontology relationship \texttt{mapsToViolationType}. It is stored as a CSV lookup rather than as a permanent database column so that the mapping remains auditable and hand-correctable.

\subsection{Data-Quality Flags}

The ontology also includes data-quality flags that the agent must respect when forming hypotheses. These flags include the low match rate between Google Local establishments and SLCHD inspection histories, the distinction between clean inspections and missing inspection histories, possible duplicate violation-code rows within an inspection, geocoding gaps, the derived nature of \texttt{slchd\_establishment\_key}, incomplete violation line capture for some scraped records, and the unverified official meaning of \texttt{violation\_codes.asterisk\_count}.

These flags are part of the reasoning design rather than implementation notes. They instruct the agent to avoid overstating certainty, and they provide explicit uncertainty states that the visualization can surface.

\section{Agent Tools and Visual Trace Construction}
\label{app:tools}

All three agent variants share the same core database and charting tools. The conditions differ only in whether additional grounding tools are available. Table~\ref{tab:tools} summarizes the tool interface.

\begin{table*}[t]
\centering
\caption{
Agent tools and availability across the three grounding conditions. V1, V2, and V3 share the database, search, inspection history, SQL, and charting tools. V2 adds regulatory retrieval over food-safety rules, while V3 adds ontology-level violation classification and structured hypothesis output. The same logged tool calls also define the visual trace used by the Reasoning view.
}
\label{tab:tools}
\scriptsize
\begin{tabularx}{\textwidth}{p{0.23\textwidth} p{0.25\textwidth} X}
\toprule
\textbf{Tool} & \textbf{Available in} & \textbf{Purpose} \\
\midrule
\texttt{get\_schema} & V1, V2, V3 & Returns queryable tables and columns when the agent is unsure what information exists in the database. \\
\texttt{get\_column\_glossary} & V1, V2, V3 & Provides column meanings when names or types are insufficient. \\
\texttt{search\_establishments} & V1, V2, V3 & Finds, filters, and ranks restaurant candidates. Its results populate the map and candidate views, so it is required for restaurant-set questions. \\
\texttt{run\_sql} & V1, V2, V3 & Executes SQL for scalar, aggregate, or non-candidate database questions. \\
\texttt{get\_inspection\_history} & V1, V2, V3 & Retrieves inspection and violation history for a specific matched establishment. \\
\texttt{plot\_chart} & V1, V2, V3 & Runs a DuckDB query and renders the result as a chart visible to the user. \\
\texttt{retrieve\_regulation} & V2, V3 & Retrieves chunks from the R392-100 and SLCHD inspection-process corpus. \\
\texttt{classify\_violation} & V3 & Maps raw violation public-health-rationale strings to ontology-level \texttt{ViolationType} concepts. \\
\texttt{HypothesisResponse} & V3 & Structured response format requiring a hypothesis, confidence, narrative answer, supporting evidence, undermining evidence, source references, and ontology concepts. \\
\bottomrule
\end{tabularx}
\end{table*}

The tool interface is also the basis of the Reasoning view. Tool calls define the timeline steps. Calls to \texttt{search\_establishments} define candidate-set changes on the map. Calls to \texttt{retrieve\_regulation} create regulatory evidence nodes. Calls to \texttt{classify\_violation} attach ontology concepts to raw violation text. In V3, the final structured response links evidence items back to the tool calls that produced them.

\section{Agent Conditions and Prompt Templates}
\label{app:prompts}

Food Critic evaluates three agents that share the same database, the same charting policy, and the same core tools. They differ only in the grounding layers made available to the agent: V1 is the database-only baseline, V2 adds regulatory retrieval, and V3 adds ontology-guided hypothesis and evidence structure.

\subsection{Shared Charting Policy}

The following charting policy applies to all three conditions. It requires the agent to produce visual output rather than text alone and to intentionally select chart forms.


\begin{promptbox}[Shared charting policy included in all three system prompts.]
Every final answer must include a visualization, not text alone --- call
plot\_chart yourself whenever search\_establishments' map isn't already the
right visual for the question. The app auto-generates a fallback chart if you
forget entirely, but call plot\_chart yourself so you can pick the chart form
that actually fits the question, rather than relying on the fallback's best
guess. Cite the actual numbers you found in your final answer, in addition to
the chart --- the chart backs up the claim, it doesn't replace stating it.

If the question asks for a ``detailed'', ``comprehensive'', ``thorough'', or
``full'' visual analysis, or itself has more than one facet (e.g.\ ``which
restaurants improved and which got worse'', ``trend over time by category''),
one chart is not enough --- call plot\_chart once per facet instead of cramming
every facet into a single combined chart. Each call should look like what a
competent visual-analytics tool would produce for that facet, not an
afterthought: pick the chart\_type that actually matches the data's shape; use
your visual-analytics judgment to include the most relevant informational
charts for each facet, give it a specific title naming what it shows, and only
include as many rows/categories as a reader could actually read (e.g.\ top N,
not all N). A ``detailed'' answer is judged by whether each chart on its own
answers one piece of the question clearly, not by how much text surrounds it.
\end{promptbox}

\subsection{\texttt{plot\_chart} Tool Instruction}

The agent has access to a charting function, \texttt{plot\_chart}, which runs a DuckDB SQL query and renders the result as a chart shown to the user. It requires a query, a \texttt{chart\_type} (one of bar, line, scatter, pie, or histogram), and x; the y, color, and title arguments are optional. x, y, and color must each be a column name present in the query's SELECT output. The agent is prompted to use this whenever the user's question calls for a visual rather than just a number.

\subsection{V1: Database-Only Agent}

V1 is the data-only baseline. It has access to the database schema, DuckDB database, structured establishment search, inspection-history lookup, and chart rendering. It has no regulatory retrieval and no ontology-guided response schema.

\begin{promptbox}[V1 system prompt]
You are a food-safety data analyst answering questions about restaurants in
Salt Lake County/Utah, grounded only in \texttt{db/food\_health.duckdb} --- no
RAG, no ontology-guided reasoning in this variant. Call \texttt{get\_schema}
first if you're unsure what's queryable, and \texttt{get\_column\_glossary} for
any column whose meaning isn't obvious from its type.

Use \texttt{search\_establishments} for ANY question whose answer is a set of
restaurants --- lookups, filters, AND rankings (``top N by X'', ``most/worst
Y''), including rankings by violation count (it has \texttt{sort\_by=}
\texttt{critical\_violation\_count} / \texttt{total\_violation\_count} /
\texttt{avg\_rating} / \texttt{num\_of\_reviews} / \texttt{inspection\_score} /
\texttt{distance\_mi}, and \texttt{sort\_desc=}). Its results are what the UI
shows on the map --- a hand-written \texttt{run\_sql} ranking of establishments
will NOT appear there, so use \texttt{search\_establishments} for that even
though \texttt{run\_sql} could technically answer it too. Reserve
\texttt{run\_sql} for pure scalar/aggregate answers that aren't about a set of
restaurants (e.g.\ ``what's the average score across all inspections''), and
\texttt{get\_inspection\_history} for a specific matched establishment's
inspection/violation record.
\end{promptbox}

\subsection{V2: Regulatory RAG Agent}

V2 includes all V1 tools plus \texttt{retrieve\_regulation}, which searches the R392-100 and SLCHD inspection-process corpus. This condition tests whether explicit access to food-safety regulation changes answers for questions about scoring, rule meaning, and public-health interpretation.

\begin{promptbox}[V2 system prompt]
You are a food-safety data analyst answering questions about restaurants in
Salt Lake County/Utah. You have two grounding sources:
\texttt{db/food\_health.duckdb} (via \texttt{search\_establishments} /
\texttt{run\_sql} / \texttt{plot\_chart} / \texttt{get\_inspection\_history})
for facts about specific restaurants, and \texttt{retrieve\_regulation} for
Utah's R392-100 Food Service Sanitation Rule and SLC county's
inspection-process page. Call \texttt{get\_schema} first if you're unsure
what's queryable, and \texttt{get\_column\_glossary} for any column whose
meaning isn't obvious.

Use \texttt{search\_establishments} for ANY question whose answer is a set of
restaurants --- lookups, filters, AND rankings (``top N by X'', ``most/worst
Y''), including rankings by violation count (it has \texttt{sort\_by=}
\texttt{critical\_violation\_count} / \texttt{total\_violation\_count} /
\texttt{avg\_rating} / \texttt{num\_of\_reviews} / \texttt{inspection\_score} /
\texttt{distance\_mi}, and \texttt{sort\_desc=}). Its results are what the UI
shows on the map --- a hand-written \texttt{run\_sql} ranking of
establishments will NOT appear there. Reserve \texttt{run\_sql} for pure
scalar/aggregate answers not about a set of restaurants.

Use \texttt{retrieve\_regulation} when the user asks why something matters,
what a rule requires, or how scoring/risk levels work, and cite what it
returns in your answer.
\end{promptbox}

\subsection{V3: Ontology-Guided Agent}

V3 includes all V2 tools plus \texttt{classify\_violation} and a structured \texttt{HypothesisResponse} output format. This condition tests whether requiring typed hypotheses and evidence makes the agent's reasoning more inspectable and easier to render as visual provenance.

\begin{promptbox}[V3 system prompt: grounding and tool use]
You are a food-safety data analyst answering questions about restaurants in
Salt Lake County/Utah. You have three grounding sources:
\texttt{db/food\_health.duckdb} (via \texttt{search\_establishments} /
\texttt{run\_sql} / \texttt{plot\_chart} / \texttt{get\_inspection\_history})
for facts about specific restaurants, \texttt{retrieve\_regulation} for Utah's
R392-100 Food Service Sanitation Rule and SLC county's inspection-process
page, and \texttt{classify\_violation} to translate a raw
\texttt{violation\_phr} string into the ontology's \texttt{ViolationType}
concept. Call \texttt{get\_schema} first if you're unsure what's queryable,
and \texttt{get\_column\_glossary} for any column whose meaning isn't obvious.

Always use \texttt{search\_establishments} (not hand-written SQL) to find
candidate restaurants for a hypothesis --- including ranking questions
(``top N by X'', ``most/worst Y''), via its \texttt{sort\_by=}
\texttt{critical\_violation\_count} / \texttt{total\_violation\_count} /
\texttt{avg\_rating} / \texttt{num\_of\_reviews} / \texttt{inspection\_score} /
\texttt{distance\_mi} and \texttt{sort\_desc=} --- every establishment it
returns becomes part of the map/evidence view shown to the user, so a
hand-written SQL query (even one that technically answers the question) would
leave the Reasoning tab empty.

\texttt{\{grounding\_policy\}}
\end{promptbox}

\begin{promptbox}[V3 system prompt: answer structure]
Your final answer must be structured, not just prose.
\texttt{\{charting\_policy\}} \texttt{search\_establishments}' results
additionally populate the Reasoning tab's map step automatically, on top of
whatever \texttt{plot\_chart} you called. After gathering facts with your
tools, decide whether the question calls for a hypothesis about an
establishment (a recommendation, a warning, or a trend claim) or is just a
plain data lookup.

\texttt{\{hypothesis\_guidance\}}
\end{promptbox}

\begin{promptbox}[V3 system prompt: evidence requirements]
For every hypothesis you form (anything other than `none'), back it with
\texttt{supporting\_evidence} and, where it exists,
\texttt{undermining\_evidence} --- each item must cite a specific
\texttt{search\_establishments} / \texttt{run\_sql} result or
\texttt{retrieve\_regulation} chunk (\texttt{source\_ref}) and name the
ontology concept it bears on (\texttt{relates\_to\_concept}), e.g.\
`\texttt{ViolationCode.critical}', `\texttt{InspectionTrend.worsening}', or
`\texttt{ViolationType.temperature\_control}' (call
\texttt{classify\_violation} to get this for a violation row).

Do not fabricate evidence --- every item must trace back to an actual tool
call you made. If the question is a plain lookup (e.g.\ ``what's the address
of X''), set \texttt{hypothesis} to `none' and leave both evidence lists
empty; still put the answer in \texttt{narrative\_answer}. Leave
\texttt{candidate\_pool} empty --- it's filled in separately from your actual
\texttt{search\_establishments} calls, not by you.
\end{promptbox}

\subsection{V3 Hypothesis Guidance}

The V3 prompt dynamically inserts hypothesis guidance from the ontology. The following vocabulary is available to the agent.

\begin{promptbox}[V3 hypothesis vocabulary]
Fixed hypothesis vocabulary (pick one of these, or `none' if the question is a plain data lookup that doesn't call for a hypothesis):

- safe\_bet: inspection\_score low (0-3) and stable / no critical ViolationCode on last 1-2 inspections / InspectionTrend stable/improving / avg\_rating >= 4.3 with meaningful num\_of\_reviews

- health\_risk\_concern: ViolationCode.critical = true on most recent inspection / ViolationType temperature\_control or cross\_contamination + critical / inspection\_score elevated on most recent visit / violation\_occurrences > 1 for the same code on one inspection

- improving\_compliance: InspectionTrend improving / critical-violation count decreasing over the window / Followup (type 2) after a prior high-score inspection, with a lower score

- declining\_compliance: InspectionTrend worsening/increasing critical-violation count or new critical codes / inspection\_score trending up over the last 3 inspections

- repeat\_critical\_violator: same critical violation\_code\_id across multiple inspection\_dates / violation\_cos = false on critical items / Critical Item (type 7) or Followup (type 2) present in history

- no\_inspection\_data\_available: Establishment.slchd\_establishment\_key is null -- no matchedToSLCHDListing exists / applies to 98.4\% of establishments; state this explicitly rather than staying silent on inspection history

- well\_reviewed: avg\_rating >= 4.5 (75th percentile) / num\_of\_reviews >= 225 (dataset median) / does not require an SLCHDListing match; usable for the 98.4\% with no inspection data

- reputation\_inspection\_mismatch: avg\_rating >= 4.5 and num\_of\_reviews >= 225 (well\_reviewed) / but ViolationCode.critical = true on the most recent inspection, or InspectionTrend worsening / only computable when matchedToSLCHDListing exists

- limited\_track\_record: num\_of\_reviews <= 18 (10th percentile) / not enough signal to support or undermine any other hypothesis with confidence

\end{promptbox}


\begin{promptbox}[V3 data-quality flags]
Data quality flags -- respect these when forming a hypothesis or citing evidence:

- Only 139/8592 establishments (1.6\%) have a matched SLCHDListing. Don't imply inspection history for the rest.

- No violation rows for an inspection means clean, not missing data.

- 173 inspections (2.5\%) have a violation\_code\_id repeated within one inspection -- likely duplicate/OCR-artifact scraping, not double citations. Per-inspection violation counts may be inflated for these.

- 71/1341 SLCHDListing rows (5.3\%) have no lat/lon after Census + Nominatim geocoding.

- slchd\_establishment\_key is a derived NAME|ADDRESS|CITY string, not a permit id -- a renamed/re-registered same location isn't deduplicated across time.

- SLCHDListing rows with source=own\_scrape (96/1341) never captured violation line items, even where real inspections had violations.

- violation\_codes.asterisk\_count correlates with critical, but its official Utah Food Code meaning was not independently verified.
\end{promptbox}

\section{Evaluation Question Set}
\label{app:questions}

The controlled evaluation uses 18 tasks totaling 21 turns: 15 single-turn questions (Q1--Q15) and three two-turn follow-up threads (A1/A2, B1/B2, C1/C2). The second turn of each thread is asked in the same conversation as the first, so it depends on the preceding turn and lets the memory-augmented variants be evaluated on short conversational threads. The script mixes database lookup, visual analysis, regulatory interpretation, restaurant-specific judgment, and preference-based recommendation tasks.

\paragraph{Single-turn questions.}
\begin{description}
    \item[Q1] Which restaurant or restaurants violated health or food codes most?
    \item[Q2] Is there any correlation between ratings and violations among the restaurants?
    \item[Q3] Give me the name of the single restaurant with the most reviews in the database, along with its full violation history.
    \item[Q4] Find me the restaurant with the most reviews that has had a violation.
    \item[Q5] What does an inspection score mean?
    \item[Q6] What is the difference between a critical and a non-critical violation, and how does each affect an inspection score?
    \item[Q7] Is the In-N-Out Burger a safe bet to eat at right now? Show detailed visual analytics to answer this.
    \item[Q8] STELLA GRILL's inspection scores have fluctuated significantly over time. Should I be concerned? Answer with very detailed visual analytics.
    \item[Q9] What does the issue category ``cold holding'' mean?
    \item[Q10] Show me the geographic spread of restaurants with at least one critical violation, on a map.
    \item[Q11] Show me detailed visual analytics and statistics of inspection scores across all establishments.
    \item[Q12] For which restaurants has the number of critical violations increased, and for which has it decreased over the last 5 years? Show me a detailed visual analysis.
    \item[Q13] Which street has the highest number of 3 star restaurants?
    \item[Q14] Between Thai and Korean restaurants, which cuisine has the higher average rating? Which one has more violations?
    \item[Q15] I'm in the mood for breakfast, but don't feel like traveling far out of Rose Park. What do you recommend? I don't like sweet breakfasts.
\end{description}

\paragraph{Follow-up threads.}
\begin{description}
    \item[A1] How many establishments have a Salt Lake County health inspection on record?
    \item[A2] Among them, can you show me the distribution and timeline of health inspections?
    \item[B1] Compare the highest rated Chinese restaurant against the most reviewed Chinese restaurant.
    \item[B2] Which do you recommend I try?
    \item[C1] How hard is it to find halal food options in Salt Lake County?
    \item[C2] If I like vegetables, which do you recommend I try?
\end{description}

\subsection{Question Types and Expected Grounding Effects}

The questions instantiate five task types.

\begin{itemize}
    \item \textbf{Database lookup and ranking.} These questions can be answered from database tools alone and should not require regulatory retrieval or ontology-guided reasoning. These are Q1, Q3, Q4, Q13, Q14, A1, and B1.

    \item \textbf{Visual analysis.} These questions require the agent to produce charts or maps in addition to textual explanation. These are Q2, Q10--Q12, and A2.

    \item \textbf{Regulatory interpretation.} These questions require an explanation of inspection scoring, violation meaning, or rule significance. These are Q5, Q6, and Q9. V2 and V3 are expected to be better suited to these questions than V1 because they can call \texttt{retrieve\_regulation}.

    \item \textbf{Restaurant-specific safety judgment.} These questions require combining inspection history, review information, trend evidence, and uncertainty. These are Q7 and Q8. V3 is expected to be most informative here because it must produce a hypothesis label and typed supporting or undermining evidence.

    \item \textbf{Preference-based recommendation.} These questions combine user preferences with restaurant metadata and, where available, inspection evidence. These are Q15, B2, C1, and C2. These tasks are intentionally ill-defined: the system should support user judgment rather than compute a universal optimum.
\end{itemize}

\section{Scoring Rubric}
\label{app:rubric}

The scoring rubric evaluates both answer correctness and inspectability of reasoning. It is aligned with the two research questions: whether grounding changes the agent's output, and whether the resulting reasoning can be represented as visual state.

\subsection{Factual Correctness}

A response is factually correct when its claims match the database records or retrieved regulation chunks returned by the tools. For database lookup, ranking, and aggregate questions, correctness is judged against the executed SQL result, structured establishment search result, or inspection-history result. Unsupported numeric claims are penalized.

\subsection{Appropriate Tool Use}

A response should use the appropriate tool for the task. Restaurant-set questions should use \texttt{search\_establishments} so that the candidates appear in the interface. Scalar and aggregate questions may use \texttt{run\_sql}. Restaurant-specific inspection questions should call \texttt{get\_inspection\_history}. Regulatory interpretation questions should call \texttt{retrieve\_regulation} in V2 and V3.

\subsection{Visualization Adequacy}

Because every final answer is expected to include a visualization, responses are scored for whether the chart or map fits the question. A chart is adequate when the chart type matches the data structure, the title identifies the analytic claim, the axes or categories are interpretable, and the number of displayed rows or categories is readable. For detailed visual-analysis prompts, a single chart may be insufficient if the question has multiple facets.

\subsection{Grounding and Abstention}

A response should acknowledge missing inspection evidence rather than implying a clean safety record. In particular, review sentiment should not be used as a substitute for inspection evidence without caveat. A response is penalized if it treats \texttt{slchd\_establishment\_key = null} as evidence of safety.

\subsection{Regulatory Grounding}

For questions about inspection scores, violation meaning, critical versus non-critical violations, or food-code interpretation, V2 and V3 responses should cite retrieved regulatory evidence. A response is penalized if it offers an unsupported rule interpretation without using the retrieval tool.

\subsection{Ontology and Evidence Structure}

For V3, the response should contain a valid hypothesis label, confidence value, narrative answer, supporting evidence, undermining evidence where applicable, source references, and ontology concept tags. Evidence items must trace back to actual tool calls and should be specific enough to render as visual marks.

\subsection{Visual Trace Completeness}

A run supports RQ2 when the logged trace contains enough information to reconstruct the Reasoning view. This requires a sequence of tool calls, candidate-set outputs from \texttt{search\_establishments}, retrieved regulation chunks where used, ontology-tagged evidence items in V3, and chart specifications where visual analytics are requested.

\section{Additional Results}
\label{app:results}

This section reports additional results from the controlled prompt runs. For each model provider and agent condition, we include aggregate success rates, predicted-winner agreement, invalid tool-call rates, invalid structured-output rates, token usage, and qualitative notes.

\subsection{Total Cost}
\begin{figure}[t]
  \centering
  \includegraphics[width=\linewidth]{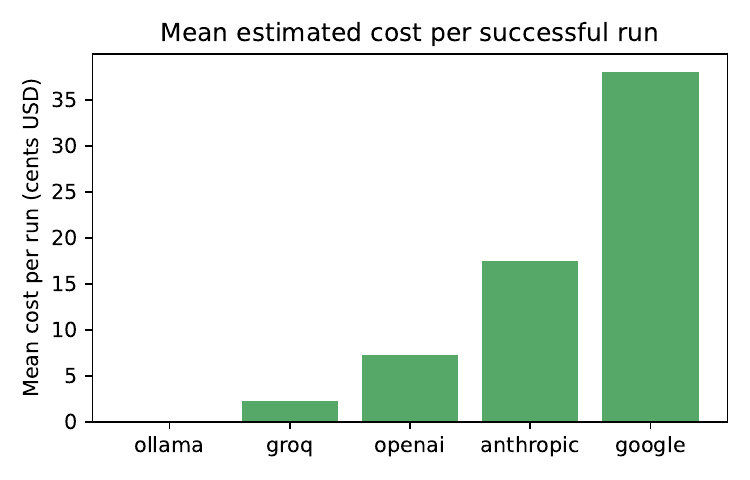}
  \caption{Mean estimated cost per successful run by model provider. Costs are normalized by successful, non-error runs so that providers can be compared despite different failure rates from rate limits, structured-output constraints, or malformed tool calls. The figure reports operational cost as part of the system-level evaluation of agentic visualization reliability.}
  \label{fig:cost}
\end{figure}

Here we summarize the total cost of running each agent variant across the four LLM providers. Anthropic's \texttt{claude-sonnet-5} was the most expensive at \$6.80, followed closely by Google's \texttt{gemini-3.6-flash} at \$6.06 and Groq's \texttt{llama-3.3-70b-versatile} at \$5.30. OpenAI's \texttt{gpt-5.6-terra} was by far the cheapest at \$1.85, roughly a quarter of the cost of the most expensive variant. Overall, the total expenditure across all four variants amounted to \$20.01. Gemma 4 26B (\texttt{gemma4:26b}) ran locally through Ollama and incurred no API cost; it is excluded from these totals and appears at \$0 in Figure~\ref{fig:cost}. See Figure \ref{fig:cost} for a comparison of mean costs per run.

\subsection{Qualitative Trace Example}
\label{sec:qualitative-trace}

\begin{figure*}[t]
  \centering
  \includegraphics[width=0.85\linewidth]{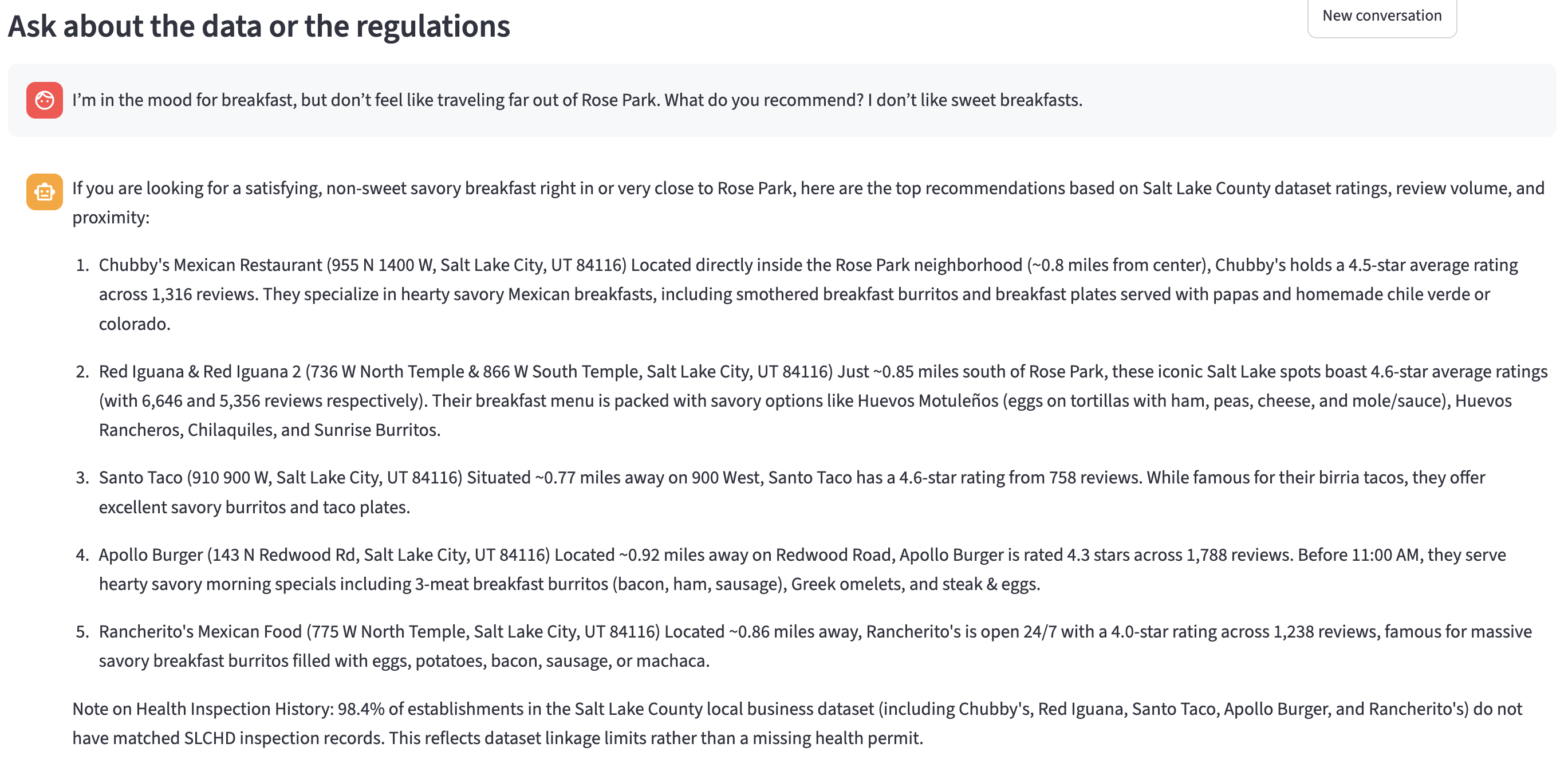}
  \caption{V3: User prompt and final response.}
  \label{fig:trace-v3-txt}
\end{figure*}

\begin{figure*}[t]
  \centering
  \includegraphics[width=0.85\linewidth]{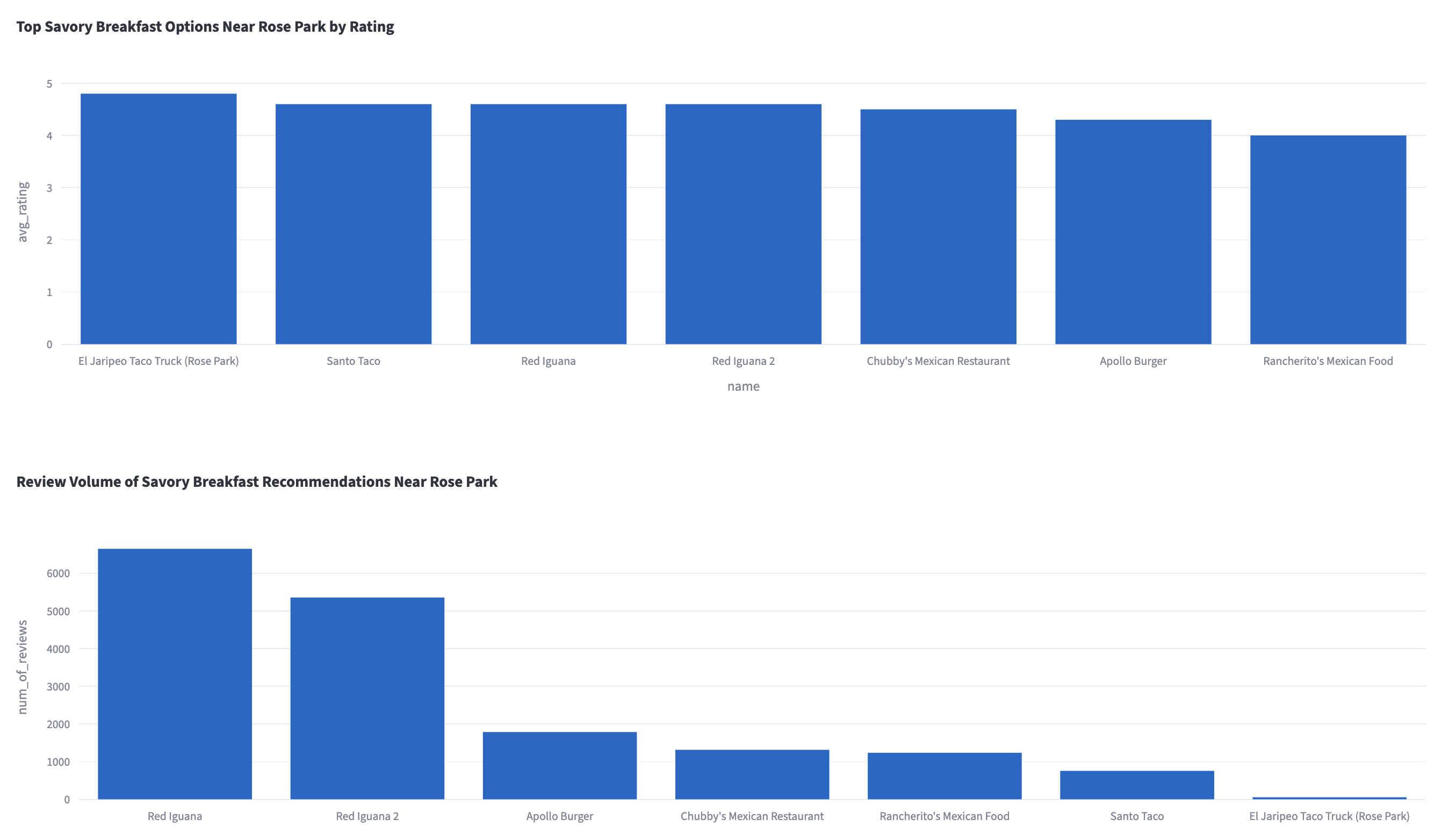}
  \caption{V3: visualization state.}
  \label{fig:trace-v3-viz}
\end{figure*}

\begin{figure*}[t]
  \centering
  \includegraphics[width=0.85\linewidth]{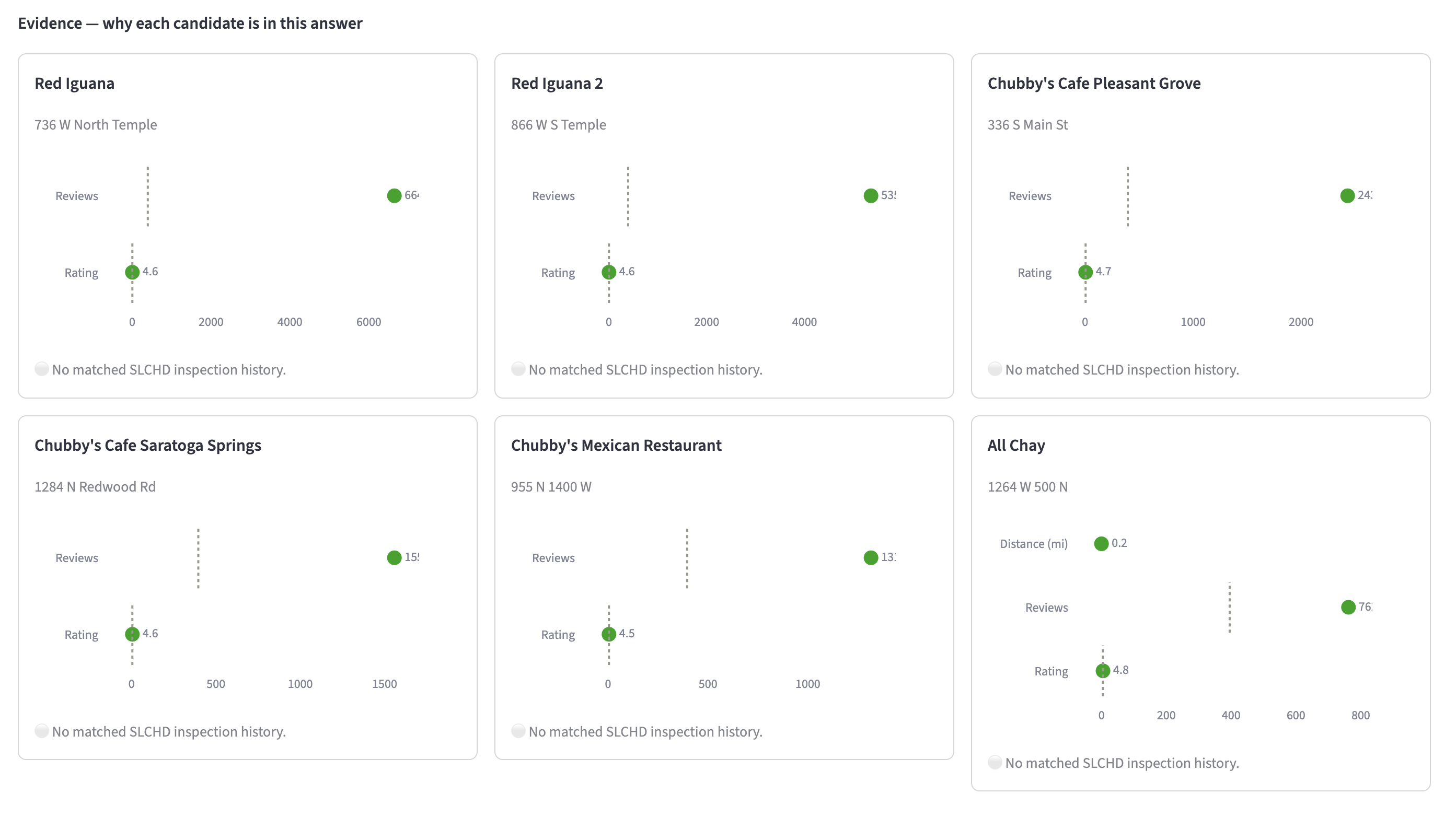}
  \caption{V3: Reasoning tab, per-candidate evidence glyphs showing reviews, rating, and
  SLCHD match status.}
  \label{fig:trace-v3-glyphs}
\end{figure*}

Figures~\ref{fig:trace-v3-txt}--\ref{fig:trace-v3-glyphs} show V3's response to the question "I’m in the mood for breakfast, but don’t feel like traveling far out of Rose Park. What do you recommend? I don’t like sweet breakfasts."

\subsection{Ontology Structured Evidence by Gemini V3}
Table \ref{tab:rq2-evidence} shows the evidence produced by Gemini with agent variant V3.

\begin{table}[t]
  \centering
  \caption{Structured evidence produced by Gemini V3 for the Stella Grill safety question. Each row is an evidence item from the ontology-guided response, labeled as supporting or undermining the final hypothesis. The table shows the evidence source type, the ontology concept used to interpret the evidence, and a concise summary of the claim. This structure makes the final answer auditable at the evidence-item level rather than only as free-form prose.}
  \label{tab:rq2-evidence}
  \small
  \begin{tabular}{@{}cp{0.13\linewidth}p{0.1\linewidth}p{0.18\linewidth}p{0.45\linewidth}@{}}
    \toprule
    \# & Evidence Direction & Source type & Concept & Summary \\
    \midrule
    1 & supports & Database & Violation Code: Critical & Equipment/food-contact surface cleaning (4.4.82, critical) cited across 8 inspection dates, 2022-2025. \\
    2 & supports & Database & Violation Type: Temperature Control & Cold holding violations (4.3.59, critical) cited across 5 dates, incl.\ temperatures as high as 56\textdegree F. \\
    3 & supports & Database & Worsening Inspection Trend & Score of 185, 10 critical violations, on 2023-07-13; resulted in a Summary Permit Suspension. \\
    4 & supports & RAG chunk & Violation Code: Critical & R392-100 \S~8-904.10 authorizes summary permit suspension when an inspection reveals an imminent health hazard. \\
    5 & undermines & Database & Improving Inspection Trend & Most recent inspection (2026-02-03) improved to score 10 with only 1 critical violation. \\
    6 & undermines & Database & Average Establishment Rating & 4.5-star public rating across 825 Google Local reviews. \\
    \bottomrule
  \end{tabular}
\end{table}

\end{document}